\documentclass[10pt,journal]{IEEEtran}

\usepackage{color}
\pdfoutput=1

\usepackage[utf8]{inputenc}
\usepackage{threeparttable}
\usepackage{multirow}
\usepackage{booktabs,tabularx}
\usepackage{hyperref}
\usepackage{url}
\usepackage{float}

\usepackage{amsmath}
\usepackage{amssymb}
\usepackage{mathtools}
\usepackage{multirow}
\usepackage{color}

\definecolor{ex1}{rgb}{0, 0.3, 1}
\definecolor{ex2}{rgb}{0, 0.8961, 0.9709}
\definecolor{ex3}{rgb}{0.4902, 1, 0.4775}
\definecolor{ex4}{rgb}{0.9709, 0.9593, 0}
\definecolor{ex5}{rgb}{1, 0.4074, 0}

\begin{document}

\title{Strengths and Weaknesses of Deep Learning Models for Face Recognition Against Image Degradations}

\author{Klemen Grm$^{1, 2, 3}$\thanks{$^1$ Correspondence email: \texttt{klemen.grm@fe.uni-lj.si}}, Vitomir Štruc$^{2, 3}$\thanks{$^2$ First authors with equal contributions}, Anais Artiges$^4$\thanks{$^3$ University of Ljubljana, Faculty of electrical Engineering}, Matthieu Caron$^4$\thanks{$^4$ ENSEA, Graduate School in Electrial and Computer Engineering}, Hazim Kemal Ekenel$^5$\thanks{$^5$ Istanbul Technical University}}

\markboth{Submitted for publication to IET Biometrics}{}

\maketitle

\begin{abstract}
Deep convolutional neural networks (CNNs) based approaches are the state-of-the-art in various computer vision tasks, including face recognition. Considerable research effort is currently being directed towards further improving deep CNNs by focusing on more powerful model architectures and better learning techniques. However, studies systematically exploring the strengths and weaknesses of existing deep models for face recognition are still relatively scarce in the literature. In this paper, we try to fill this gap and study the effects of different covariates on the verification performance of four recent deep CNN models using the Labeled Faces in the Wild (LFW) dataset. Specifically, we investigate the influence of covariates related to: \textit{image quality} -- blur, JPEG compression, occlusion, noise, image brightness, contrast, missing pixels; and \textit{model characteristics} -- CNN architecture, color information, descriptor computation; and analyze their impact on the face verification performance of AlexNet, VGG-Face, GoogLeNet, and SqueezeNet. Based on comprehensive and rigorous  experimentation, we identify the strengths and weaknesses of the deep learning models, and present key areas for potential future research. Our results indicate that high levels of noise, blur, missing pixels, and brightness have a detrimental effect on the verification performance of all models, whereas the impact of contrast changes and compression artifacts is limited. It has been found that the descriptor computation strategy and color information does not have a significant influence on performance.
\end{abstract}

\section{Introduction}

Recent advances in deep learning and convolutional neural networks (CNNs) have contributed to significant performance improvements in a number of computer vision problems, ranging from low-level vision tasks, such as saliency detection and modeling~\cite{Bruce_2016_CVPR, Li_2016_CVPR} to higher-level problems such as object detection~\cite{gidaris2015object,ren2015faster},  recognition~\cite{girshick2014rich, Gidaris_2016_CVPR, Liu_2016_CVPR, He_2016_CVPR,szegedy2015going}, tracking~\cite{Alahi_2016_CVPR, wang2013learning, Wang_2016_CVPR}, or semantic segmentation~\cite{badrinarayanan2015segnet, chen2014semantic, sharma2015deep}. Deep learning-based approaches have been particularly successful in the field of face recognition, where contemporary deep models now report near perfect performance on popular, long-standing benchmarks such as Labeled Faces in the Wild~\cite{huang2007labeled}, which due to its difficulty, represented the \textit{de facto} standard for evaluating face recognition technology for nearly a decade. %

Most of the ongoing research on deep learning-based face recognition focuses on new model architectures, better techniques for exploiting the generated face representations, and related approaches aimed at improving both the performance and robustness of deep face recognition technology on common benchmark tasks~\cite{CVPRDeepFace, facenet, VGGface}. Research in these areas is typically conducted on unconstrained datasets with various sources of image variability present at once, which makes it difficult to draw clear conclusions about the sources of errors and problems that are not addressed appropriately by the existing deep CNN models. Much less work is devoted to the systematical assessment of the robustness of deep learning models for face recognition against specific variations. Considering the widespread use of deep CNN models for face recognition, it is of paramount importance that the behavior and characteristics of these models are well understood and open problems pertaining to the technology are clearly articulated. 

In this paper, we contribute towards a better understanding of deep learning-based face recognition models by studying the impact of image-quality and model-related characteristics on face verification performance. We use four state-of-the-art deep CNN models, i.e., AlexNet~\cite{alexnet}, VGG-Face~\cite{VGGface}, GoogLeNet~\cite{szegedy2016rethinking}, and SqueezeNet~\cite{iandola2016squeezenet}, to compute image descriptors from input images and investigate how quality-related factors such as blur, compression artifacts, noise, brightness, contrast, and missing data affect their performance. Furthermore, we also explore the importance of color information and descriptor computation strategies through rigorous experimentation using the Labeled Faces in the Wild (LFW) benchmark~\cite{huang2007labeled}. The deep CNN models considered in this work are representatives of the most commonly employed CNN architectures in use today and were selected due to their popularity within the research community. The studied covariates, on the other hand, represent factors commonly encountered in real life that are known to affect face recognition technology to a significant extent~\cite{jain2011handbook} and have not yet been studied sufficiently in the literature in the context of deep learning.

The comprehensive analysis presented in this paper builds on the previous works from~\cite{Ghazi_2016_CVPR_Workshops, karahan2016image}. These works both focused on closed-set face identification and investigated the robustness of deep CNN models under facial appearance variations caused by head pose, illumination, occlusion, misalignment in~\cite{Ghazi_2016_CVPR_Workshops} and by image degradations in~\cite{karahan2016image}. Complementing and extending these previous works, we  provide in this paper a rigorous and systematical evaluation of the impact of various image- and model-related factors on deep learning-based face verification performance. The goal of this work is to provide answers to essential research questions, such as: Are good quality images a must for high verification performance? To what extent does image quality affect the image descriptors generated by contemporary deep models? Are certain model architectures more robust than others against variations of specific covariates? Changes in which quality characteristics are most detrimental to the verification performance? How should image descriptors be computed? Answers to these and similar questions are in our opinion crucial for a better understanding of deep learning-based face recognition technology and may point to open problems that need to be addressed in the future. In summary, we make the following contributions in this paper:

\begin{itemize}
	\item We study and empirically evaluate the effect of image quality (blur, JPEG compression, noise, contrast, brightness, missing data) and model related (color information, descriptor computation) characteristics on the face verification performance of four state-of-the-art deep CNN models on the LFW dataset.
	\item We conduct a comprehensive analysis of the experimental results, identify the most detrimental covariates affecting deep CNN models in face verification task and point to potential areas for improvement.
    \item We provide a comparative evaluation of the four deep CNN models, namely, AlexNet~\cite{alexnet}, VGG-Face~\cite{VGGface}, GoogLeNet~\cite{szegedy2016rethinking}, and SqueezeNet~\cite{iandola2016squeezenet}, and make the trained models publicly available to the research community through: {\small \url{https://github.com/kgrm/face-recog-eval}}.  
\end{itemize}

The rest of the paper is organized as follows: In section \ref{SubSec: Related}, we briefly review previous works relevant to our study. In section \ref{Sec: Methodology}, we describe the evaluation methodology, models, datasets, and experimental procedures used. In section \ref{Sec: Exper}, we present quantitative results and discuss our experiments. Finally, section \ref{Sec: conclusion} concludes the paper.

\section{Related work}\label{SubSec: Related}

Understanding the strengths and weaknesses of machine learning models is of paramount importance for real-world applications and a prerequisite for identifying future research and developments needs. Papers on the analysis of deep models appear in the literature in either \textit{i)} work that focuses specifically on the characteristics of deep models,
or \textit{ii)} work that explores the characteristics of deep models as part of another contribution.
Papers from the first group, such as ours, typically explore various models and as the main contribution present general findings that apply to several deep models,
while papers from the second group propose a new deep learning approach and then analyze its characteristics. Both groups of work typically contribute to a better understanding of deep models, but differ in their generality, i.e., the number of models the findings apply to.

An example of work studying the impact of various image-quality covariates on the performance of several deep CNN models was presented by Dodge and Karam in~\cite{dodge2016understanding}. Here, the authors explored the influence of noise, blur, contrast, and JPEG compression on the performance of four deep neural network models applied to the general image classification task. The authors concluded that noise and blur are the most detrimental factors.

In~\cite{Devil-chatfield2014return} Chatfield et al. compared traditional machine learning models and deep learning models on equal footing by using the same data augmentation and preprocessing techniques that are commonly used with convolutional neural networks on traditional machine learning models. The authors also explored the importance of color information, but focused on the impact of color on traditional models rather that on its role in deep learning. The main finding of this work was that deep learning models have an edge over traditional machine learning models. However, data augmentation, color information, and other preprocessing tasks were found to be important, as these approaches also helped to improve the performance of traditional machine learning models.

An alternative view on covariate analyses involving deep models was recently presented by Richard-Webster et al. in~\cite{richardwebster2016psyphy}. In this work the authors compare and evaluate several deep convolutional neural network architectures from the perspective of visual psychophysics. In the context of the object recognition task, they use procedurally rendered images of 3-D models of objects corresponding to the ImageNet object classes to determine the ``canonical views'' learned by deep convolutional neural networks and determine the networks' performance when viewing the objects from different angles and distances or when the images are subjected to deformations such as random linear occlusion of the object bounding box, Gaussian blur, and brightness changes. The main point made by the authors is that model comparison must be conducted under variations of the input data, or in other words, the analysis of the robustness of the models should be used as a methodological tool for model comparison. 

Our work builds on the preliminary results reported in~\cite{karahan2016image} and~\cite{Ghazi_2016_CVPR_Workshops} and extends our previous results to face verification experiments on the LFW dataset and a wide range of image-quality and model-related covariates. The analysis includes a larger number of deep CNN models and is significantly more comprehensive in terms of amount of analyzed factors. 

Dosovitskiy et al. describe research belonging to the group of model-analysis work in~\cite{dosovitskiy2014discriminative}. Here, the authors present an evaluation of the performance of their convolutional neural network in the presence of image transformations and deformations in the context of unsupervised image representation learning.

They conclude that combining several sources of image transformations can allow convolutional neural networks to better learn general image representations in an unsupervised manner. Similar to this work, we study in this paper the effects of image deformations on the learned image representations. However different from~\cite{dosovitskiy2014discriminative}, we assess several convolutional neural networks trained in a supervised manner. 

Another work from this group was presented by Zeiler and Fergus in~\cite{zeiler2014visualizing}. Here, the authors studied the effects of image covariates including rotation, translation, and scale in the context of interpreting and understanding the internal representations produced by deep convolutional neural networks trained on the ImageNet object classification task. In their experiments, the invariance of their convolutional neural network to the studied covariates was found to increase significantly with network depth. They also found the deep neural network features to increase in discriminative power with network depth in the context of transfer learning.

More recently, Lenc and Vedaldi in~\cite{lenc2015understanding} evaluates how well the properties of equivariance, invariance, and equivalence are preserved in the presence of image transformations by various image representation models including deep convolutional neural networks. The transformations studied include rotation, mirroring,
and affine transformations of the input images. Amongst their findings, representations based on deep convolutional neural networks were found to be better than other studied representations at learning either invariance or equivariance to the studied transformations based on training objectives.

\section{Methodology}\label{Sec: Methodology}
In this section, we first explain the evaluation methodology and introduce the four deep CNN models selected for the analysis. We then proceed by presenting the dataset and procedure used to train the deep models and conclude the section with a detailed description of the covariates considered in this work.

\subsection{Evaluation methodology}\label{SubSec: Methodology}

To assess the robustness of deep CNN models against various image degradations in face verification, we take four pretrained state-of-the-art deep models and use the feature output of each model as the image descriptor of the given input face image, i.e.:
\begin{equation}
\mathbf{y} = f(\mathbf{x}),
\label{feature_extraction}
\end{equation}
where $\mathbf{x}\in\mathbb{R}^d$ denotes the input image, $f(\cdot)$ represents the selected deep model and $\mathbf{y}\in\mathbb{R}^{d'}$ stands for the computed image descriptor. The dimensionality of the image descriptor, $d'$, varies from model to model and depends on the design choices made during network construction. Once the descriptors are computed for a pair of face images, a similarity score is calculated based on the cosine similarity between the two descriptors and used to make a verification decision:   
\begin{equation}
g(\mathbf{x}_1, \mathbf{x}_2, f, T)=\left\{\begin{matrix*}[l]
w_1,& \text{if}\ \delta(f(\mathbf{x}_1), f(\mathbf{x}_2))=\delta(\mathbf{y}_1,\mathbf{y}_2) \geq T \\ 
w_2,& \text{otherwise}
\end{matrix*}\right.
\label{verification_eqn}
\end{equation}
where $\mathbf{x}_1$ and $\mathbf{x}_2$ are the input images, $\delta(\cdot, \cdot)$ is the cosine similarity, $T$ is a predefined decision threshold, and $w_1$ and $w_2$ represent classes of matching and non-matching identities, respectively. Thus, a pair of images should be classified into the class $w_1$ if the input images belong to the same identity and into the class $w_2$ if not.
To assess the robustness of the deep models with respect to different image-quality covariates, we artificially degrade one of the images in Eq.~(\ref{verification_eqn}) by adding different levels of noise, blur, compression artifacts and the like and leave the second one unaltered. With this procedure, we are able to directly observe the change in verification performance as a consequence of the change in image quality and establish a connection between a given image-quality aspect and the performance of the deep model.

We report our results using the performance metrics introduced by the LFW verification protocol~\cite{lfw_protocol}, namely, the mean and standard deviation of the verification accuracy under a $10$-fold cross-validation experimental protocol. As prescribed by the LFW experimental protocol, the decision threshold $T$ is selected separately for each fold.

\subsection{Deep CNN Models}\label{SubSec: DeepM}

We consider four recent deep CNN models in our experiments that are representative of the most popular network architectures commonly used for recognition problems, i.e.:
\begin{table*}[t]
\renewcommand{\arraystretch}{1.2}
\centering
\caption{Comparison of the quantitative properties of the deep learning models considered.}
\label{dnn_properties}
\begin{tabular}{lrrrrr} \hline
Model & \#parameters & input size & output size & \#layers & FLOPS / fwd. pass \\ \hline
AlexNet~\cite{alexnet} & $58\,282\,752$  & $(3, 224, 224)$ & $4096$ & $7$ & $1.1\times 10^9$\\ 
VGG-Face~\cite{VGGface} & $117\,479\,232$ & $(3, 224, 224)$ & $4096$ & $15$ & $1.5\times 10^{10}$ \\ 
GoogLeNet~\cite{szegedy2016rethinking} & $21\,577\,728$ & $(3, 299, 299)$ & $2048$ & $37$ & $5.6\times 10^9$ \\ 
SqueezeNet~\cite{iandola2016squeezenet} & $3\,753\,856$ &$(3, 224, 224)$ & $2048$ & $12$ & $9.7\times 10^8$\\ \hline
\end{tabular}
\end{table*}

\textbf{AlexNet}: The first model used in our evaluation is the AlexNet~\cite{alexnet}, which was the first deep convolutional neural network to successfully demonstrate performance outperforming the classical image object recognition procedures. The model consists of five sequentially connected convolutional layers of decreasing filter size, followed by three fully-connected layers. One of the main characteristics of AlexNet is the very rapid downsampling of the intermediate representations through strided convolutions and max-pooling layers. The last convolutional map is reshaped into a vector and treated as an input to a sequence of two fully-connected layers of $4096$ units in size. The output of this layer represents the image descriptor produced by AlexNet.

\textbf{VGG-Face}: The second model used in our experiments is the 16-layer VGG-Face network, initially introduced in~\cite{VGGface}.
The model has a deeper convolutional architecture than AlexNet and exploits a series of convolutional layers with small filter sizes, i.e. $3\times 3$.  Each series of convolutional layers is followed by a max-pooling layer, except for the last one, which is followed by two fully-connected layers identical to AlexNet. The output of the last fully-connected layer represents the VGG image descriptor.

\textbf{GoogLeNet}: Our third model is the GoogLeNet network, which builds on the so-called Inception architecture~\cite{szegedy2015going, szegedy2016rethinking}. Here, we use the third version of the GoogLeNet model, that is, Inception V3~\cite{szegedy2016rethinking}, which consists of a hierarchy of complex \textit{inception} modules/blocks that combine channel re-projection, spatial convolution, and pooling operations over different scales in each of the modules. The model reduces the parameter space by decomposing spatial convolutions with larger filter sizes ($n\times n$) into a sequence of two convolutional operations with respective filter sizes of $n\times 1$ and $1\times n$. The resulting network model is deeper and more complex than AlexNet or VGG-Face, but still has fewer parameters and lower computational complexity than VGG-Face. Unlike other models considered in this work, no fully-connected layers are used in GoogLeNet. Instead, the last convolutional map is subjected to channel-wise global average pooling, and the average activation values of each of the $2048$ channels are used as the feature vector of the input image.

\textbf{SqueezeNet}: The last model we assess in our experiments is a variant of the SqueezeNet network from~\cite{iandola2016squeezenet}. The network features extreme reductions in parameter space and computational complexity via channel-projection bottlenecks (or squeeze layers), and uses identity-mapping shortcut connections, similar to residual networks~\cite{He_2016_CVPR}, which allow for stable training of deeper network models. SqueezeNet was demonstrated to achieve comparable performance to AlexNet~\cite{alexnet} on the ImageNet large-scale recognition benchmark with substantial reductions in model complexity and parameter space size. The model is comprised of so-called ``fire modules'', in which the input map is first fed through a bottlenecking channel-projection layer and then divided into two channel sets. The first one is expanded through a $3\times 3$ convolution and the other through channel projection. The final convolution map is globally average-pooled into a $512$-vector and then fed to a fully-connected layer with $2048$ units. The output of this last layer is the SqueezeNet image descriptor used in our experiments.\\

Note that using deep models as ``black-box'' feature extractors is a standard way of computing (learned) descriptors from input images, as evidenced by the large body of existing research on this topic,  e.g.,~\cite{sharif2014cnn, richardwebster2016psyphy, chaib2017deep}. Furthermore, using distance metrics in the feature-space for similarity score calculation is also a standard practice in the field of biometric verification, see for example \cite{CVPRDeepFace,VGGface}. All in all the deep neural network architectures considered in this work are amongst the most popular ones found in the literature and differ greatly in computational complexity, the number of parameters, depth, and representational power. We summarize their key properties including the output feature vector (descriptor) size in Table~\ref{dnn_properties}.

\subsection{Datasets}\label{SubSec: Data}

We use separate datasets for training and evaluation.
We chose the VGG face dataset~\cite{VGGface} to train our models and the LFW dataset~\cite{huang2007labeled} to evaluate their performance. 
 
The VGG face dataset was collected during the work on the VGG-Face model~\cite{VGGface} and, as reported by the authors, comprises around $2.6\times 10^6$ images of $2622$ identities. Using the image URLs and face region coordinates published by the authors, we are able to retrieve approximately $1.8\times 10^6$ of the total $2.6\times 10^6$ face images for our version of the dataset. The structure of the VGG dataset, with a uniformly distributed and relatively large number of images per subject, $1000$, makes it similar in utility for training deep neural networks to the ImageNet dataset, which is used for image classification~\cite{russakovsky2014imagenet}.

For the experiments, we train the four deep CNN models described in the previous section from scratch using our version of the VGG face dataset to attain a fair comparison of their expressivity and other properties given the same training dataset. We train the models by appending a fully-connected softmax layer on top of each network and optimizing the model weights in accordance with the recognition performance on the VGG data. We use the Adam~\cite{kingma2015adam} gradient optimization method with the categorical cross-entropy loss function.     
During training, we randomly select $10\%$ of the images of each subject for a hold-out validation set to gauge generalization performance. Each model is trained to convergence using a GTX Titan X GPU. The training takes approximately two days each for the AlexNet and SqueezeNet models, and one week each for the GoogLeNet and VGG-FACE models.

For testing purposes, we select the Labeled Faces in the Wild~\cite{huang2007labeled} dataset, which is among the most popular datasets used to evaluate face recognition models. The dataset consists of $13233$ images of $5749$ distinct subjects, and ships with predefined training and evaluation protocols. Images of the dataset were gathered from the web and feature considerable variation in pose, lighting condition, facial expression, and background. We evaluate our models in accordance with the so-called \textit{outside-data verification protocol}, which consists of $6000$ image pairs drawn from the dataset equally divided between genuine and impostor pairs, and further equally divided into $10$ folds for cross-validation. The protocol also allows to use images not part of LFW to train the models being evaluated. 

\subsection{Performance Covariates}\label{SubSec: Covariates}

The performance of deep face recognition models depends on several factors (or covariates) that can be grouped into different categories. In this paper we are interested in factors that relate to \textit{i)} the quality of the input images
(image-quality covariates), and \textit{ii)} the characteristics of the deep models
(model-related covariates).

\textbf{Image-quality covariates:} \label{SubSec: imgCovsDesc}
To evaluate the impact of reduced image quality on the performance of our deep models, we apply image distortions of different levels/intensities to the probe images used in our verification experiments. Specifically, we consider the following:  
\begin{itemize}
  \item \textbf{Blur:} We simulate blurring effects by applying Gaussian filters  with different standard deviations $\sigma$ to the probe images. We set the filter size in accordance with the selected standard deviations, i.e., $w,h = 2\left\lceil 2\sigma \right \rceil + 1$, where $\lceil \cdot \rceil$ is the ceiling operation and $w$ and $h$ stand for the filter width and height, respectively. We vary the value of $\sigma$ from $2$ to $20$ and, thus, generate $19$ probe sets of different blur levels to investigate the impact of blurring on the performance of our deep models. 
\item \textbf{Compression:} We introduce compression artifacts by encoding the probe images with the JPEG algorithm at different quality presets. Lower quality presets correspond to more aggressive quantization of the DCT (Discrete Cosine Transform) coefficients. At the extreme, the quality of $1$ corresponds to the setting where all AC components of every MCU (Minimum Coded Unit) block are zeroed out, and each $8\times 8$ pixel block is represented by a constant color. We generate modified probe sets at quality presets of $1, 3, 5, 10, 15, 20, 25, 30, 35$ and $40$ for exploring the impact of JPEG compression. 
\item \textbf{Gaussian noise:} To study the impact of noise on the recognition performance of our deep models, we add additive Gaussian noise with a mean of 0 and various standard deviations $\sigma$ to our probe images. The modified pixel intensities are clipped to the valid dynamic range of $\left[0, 255\right]$. We generate $10$ modified probe sets for $\sigma$ values between $20 $ and $200$, with a uniform step size of $20$.
  \item \textbf{Salt-and-pepper noise:} Besides Gaussian noise, we also consider salt-and-pepper noise. Here, we truncate all color components of each image pixel to zero with a probability of $\frac{p}{2}$ and, similarly, set them to $255$ with a probability of $\frac{p}{2}$. We generate $25$ modified probe sets for probabilities $p$ between $0.02$ and $0.5$, with a uniform step size of $0.02$.
  \item \textbf{Brightness:} We simulate overexposure effects by changing the brightness level of the probe images. To this end, we  multiply the pixel intensities by a brightness factor and clip the resulting pixel values to the valid dynamic range between $\left[0, 255\right]$. We observe the impact of brightness factors between $1.5$ and $9$ with a constant step size of $0.5$ and generate $16$ probe sets for our brightness related experiments.
  \item \textbf{Contrast:} To explore the impact of contrast on the verification performance, we first subtract the central value of the dynamic range from all images. The centered images are then multiplied by a contrast factor and the offset, i.e., the central value, is added back to the image. We evaluate the performance of the models at $15$ different contrast factors between $0.03$ and $0.79$.
  \item \textbf{Missing data:} We simulate missing data (or pixels) by removing contiguous pixel areas from the probe images. Because we set all pixels in the given area to zero, the simulation of missing data is similar in effect to (artificial) partial occlusions of the face. We generate $5$ degraded probe sets with pixels missing around the mouth, nose, periocular and eye regions. To be able to remove image regions belonging to prominent facial features in a consistent manner, we use the facial landmark detection approach proposed by Kazemi and Sullivan in~\cite{kazemi2014one}.
\end{itemize}
\begin{figure}[t]
\centering{\includegraphics[width=0.99\columnwidth]{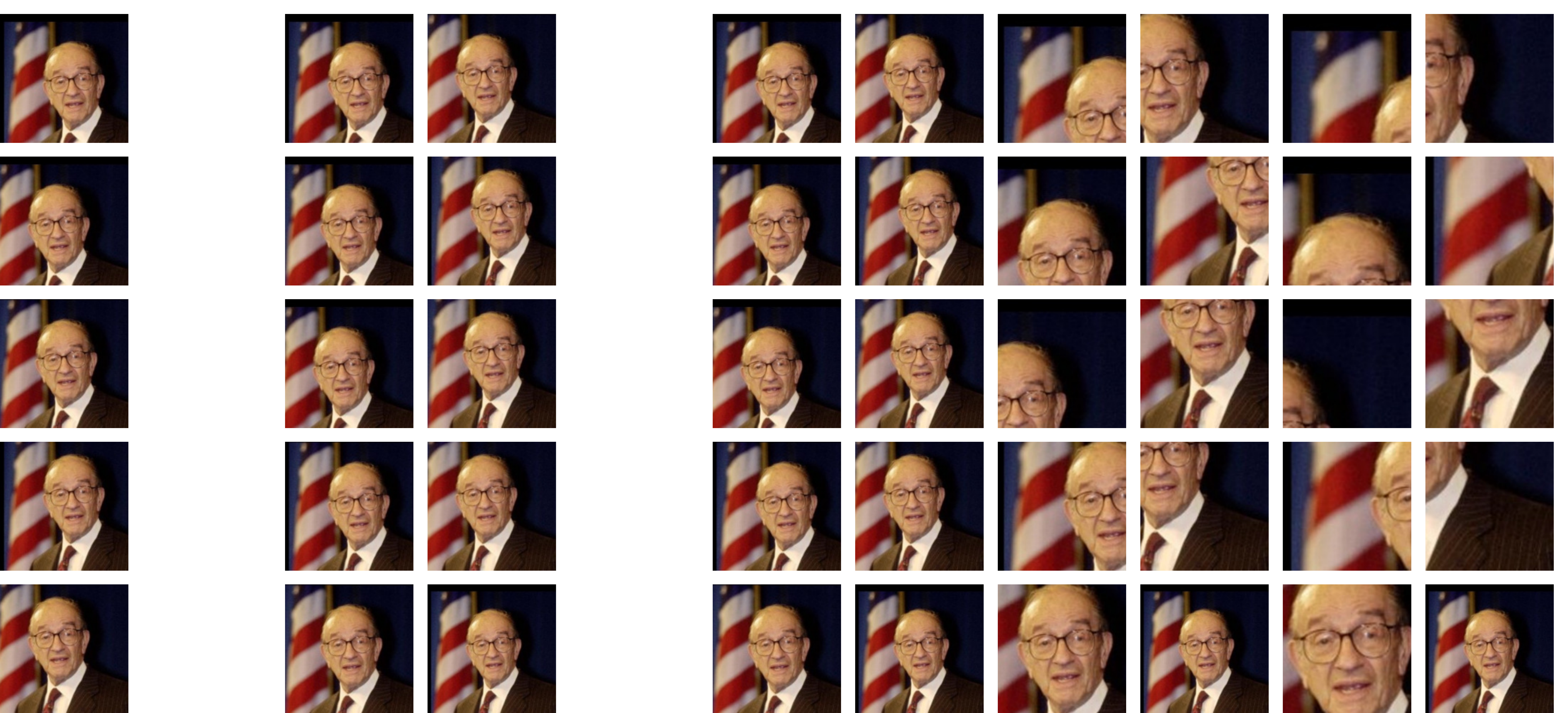}}
\caption{Illustration of the three sampling schemes used to study different descriptor-computation strategies (from left to right): the $5$-patch (left), $10$-patch (center) and $30$-patch (right) schemes.}
\label{subcrop-grids}
\end{figure}

\textbf{Model-related covariates:}
Among the model-related covariates, we explore the following ones:
\begin{itemize}
\item \textbf{Model architecture:} Arguably the most important factor affecting the performance of the existing deep face recognition approaches is the architecture of the models and corresponding training procedure used to learn the model parameters. As indicated in the introduction section, a significant amount of todays research effort related to deep models is, therefore, directed towards this area (see e.g. \cite{he2015deep,szegedy2016rethinking,iandola2016squeezenet}). In the experimental section, we account for different architectures by evaluating the four deep CNN models described in Section~\ref{SubSec: DeepM}.
  \item \textbf{Descriptor computation:} One of the key components of state-of-the-art face recognition systems is the visual descriptor used to encode the input images~\cite{Devil-chatfield2014return}. With deep learning approaches, the visual descriptor is typically computed directly from the image area returned by the face detector. The predominant approach here is to feed the detected facial area to the trained deep model and use the output of one of the top fully-connected layers as the visual descriptor of the input image. An alternative approach is to sample patches from the input image and to combine the corresponding patch representations into the final visual descriptor. Examples of the latter approach include averaging~\cite{VGGface} or stacking~\cite{sun2014deep} of patch representations and variants of Fisher Vector (FV) encoding~\cite{Chen2015Unconstrained}. For our experiments, we consider four descriptor computation strategies. The first is a simple approach, where the visual descriptor is computed directly from the facial area found by the face detector. The remaining three approaches are more complex and sample smaller patches from the facial area before averaging the patch representations generated by the models to produce the final image descriptor. We explore three sampling schemes using, $5$, $10$, and $30$ patches sampled from the detected facial area. The sampling schemes were implemented based on the suggestions in~\cite{VGGface} and are illustrated in Fig.~\ref{subcrop-grids}
  
  \item \textbf{Color space:} We consider three distinct scenarios relating to the color information of the target and probe images used in the verification experiments. In the first two cases, given a color target image, we evaluate the difference in verification performance of the deep models given either color or gray-scale probe images. In the third case, we  evaluate the performance of the models when target and query images are both gray-scale. The goal of the color-related experiments is to investigate the need for color input images and the capabilities of the models to efficiently handle gray-scale images.

\end{itemize}

\section{Experimental Results and Discussion}\label{Sec: Exper}

In this section, we describe our experiments aimed at assessing the strengths and weaknesses of the selected four deep models. We first present experiments related to image-quality covariates and then report results pertaining to the model-related covariates described in the previous section.

\subsection{Impact of image-quality covariates}\label{SubSec: image-covs}

\begin{figure*}[tb]\centering
\begin{minipage}{0.9\textwidth}
\begin{minipage}{0.47\textwidth}
\centering
\includegraphics[width=1\textwidth]{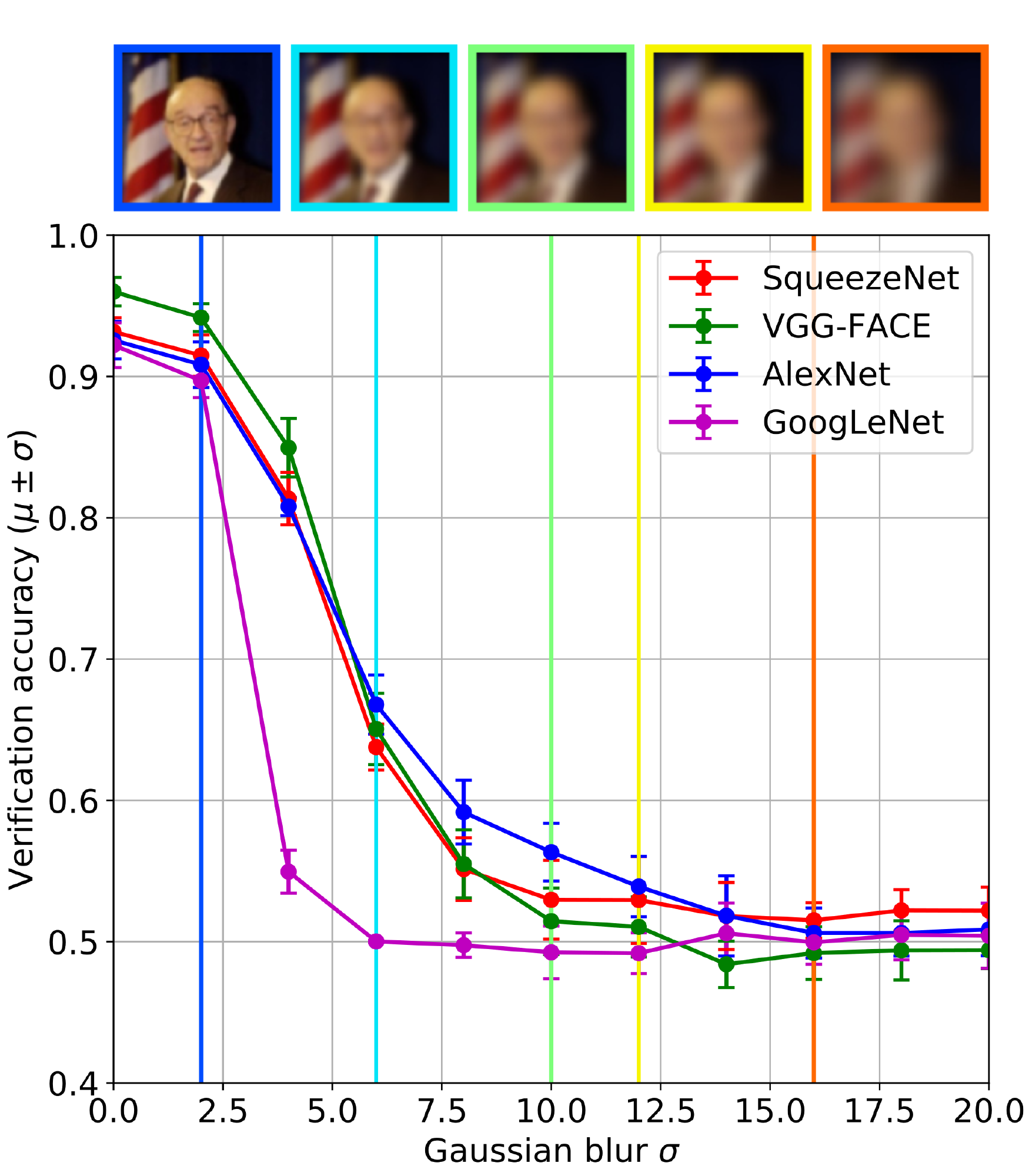}
\end{minipage}
\hfill
\begin{minipage}{0.47\textwidth}
\centering
\includegraphics[width=1\textwidth]{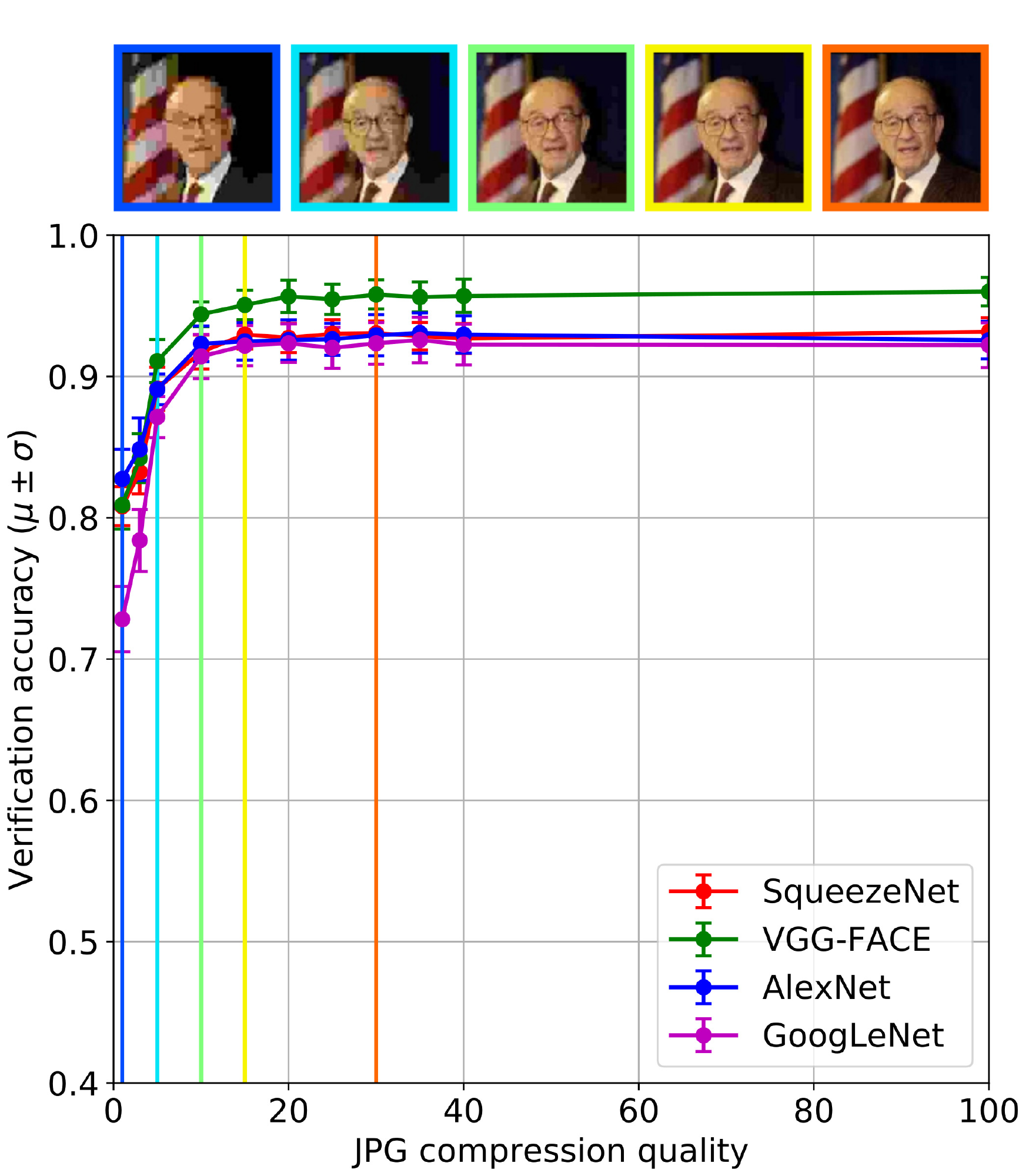}
\end{minipage}
\end{minipage}
\caption{Impact of blurring (left) and JPEG compression (right) on the performance of the four deep models. The graphs show the mean and standard deviation of the verification accuracy on the LFW dataset computed over ten folds. The images on top of the graphs show sample images generated with different levels of image distortions. The results are best viewed in color.}
\label{LFW_crop_jpg}
\end{figure*}

\begin{figure*}[t] \centering
\begin{minipage}{0.9\textwidth}
\begin{minipage}{0.47\textwidth}
\centering
\includegraphics[width=1\textwidth]{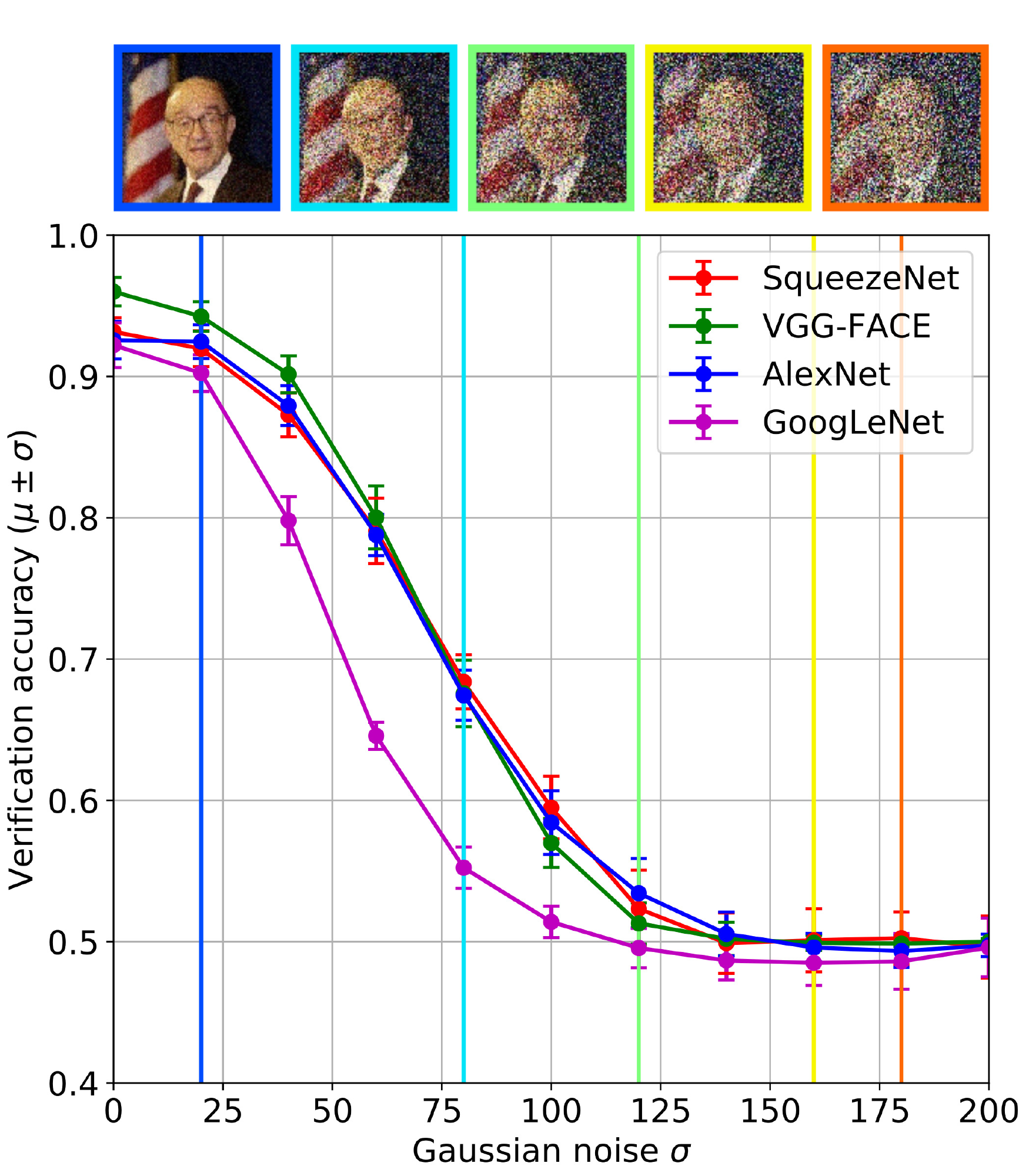}
\end{minipage}
\hfill
\begin{minipage}{0.47\textwidth}
\centering
\includegraphics[width=1\textwidth]{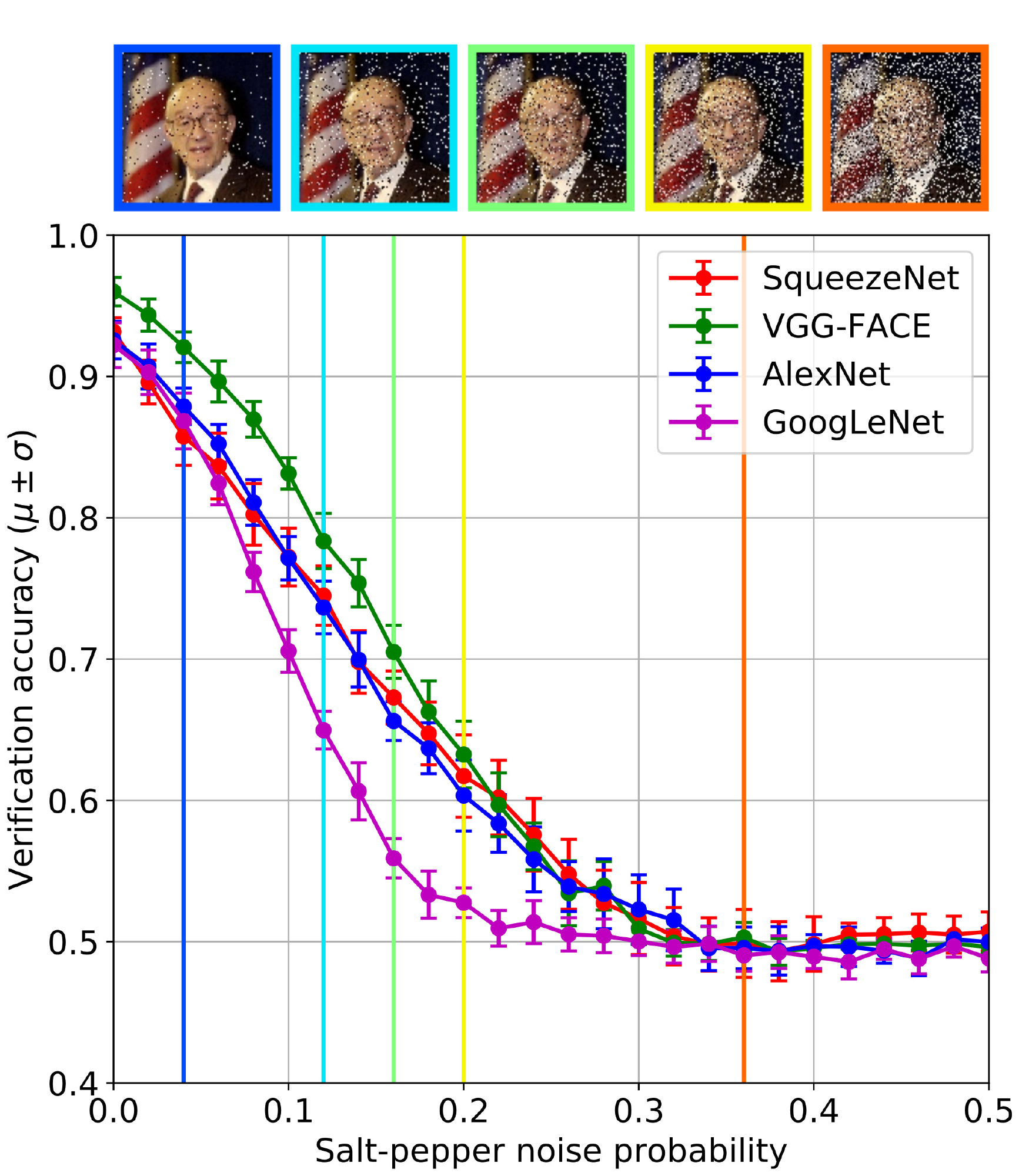}
\end{minipage}
\end{minipage}
\caption{Impact of Gaussian (left) and salt-and-pepper (right) noise on the performance of the four deep models. The graphs show the mean and standard deviation of the verification accuracy on the LFW dataset computed over ten folds. The results are best viewed in color.}
\label{LFW_gaussian_blur_noise}
\end{figure*}
\begin{figure*}[t] \centering
\begin{minipage}{0.9\textwidth}
\begin{minipage}{0.47\textwidth}
\centering
\includegraphics[width=1\textwidth]{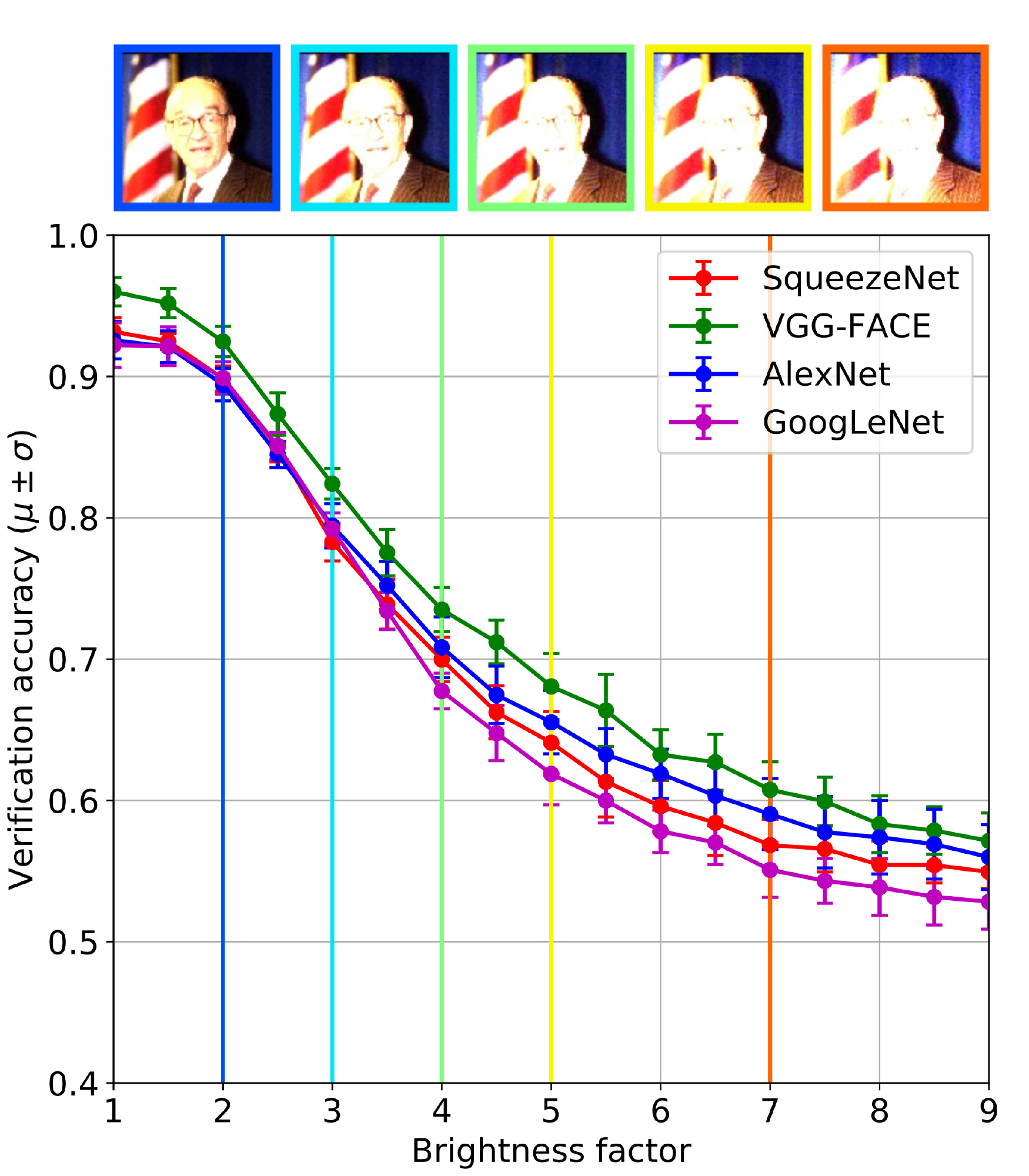}
\end{minipage}
\hfill
\begin{minipage}{0.47\textwidth}
\centering
\includegraphics[width=1\textwidth]{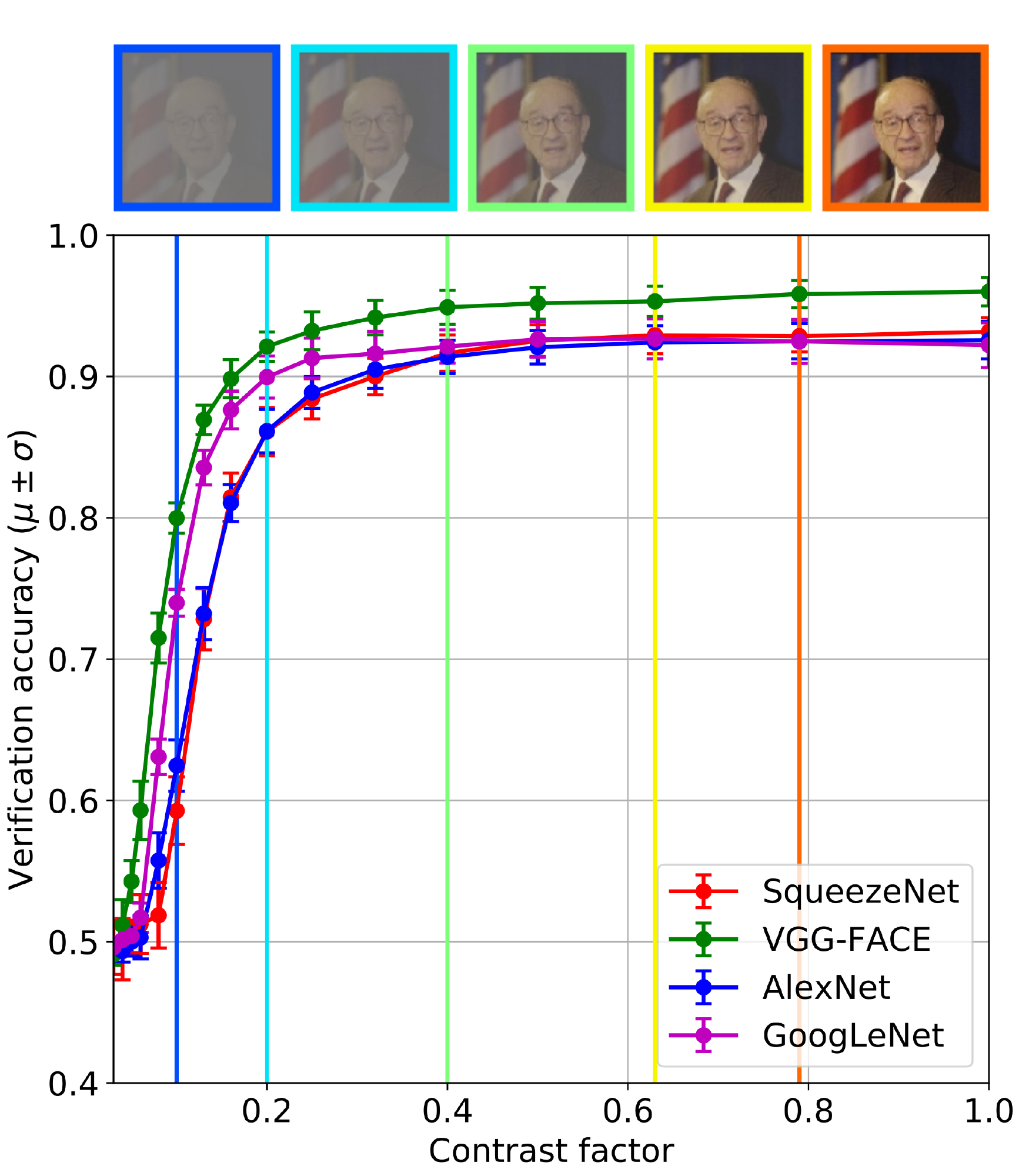}
\end{minipage}
\end{minipage}
\caption{Impact of image brightness (left) and image contrast (right) of the verification performance of our four deep models. The graphs show the mean and standard deviation of the verification accuracy on the LFW dataset computed over ten folds. The results are best viewed in color.}
\label{LFW_brightness_contrast}
\end{figure*}
\begin{figure*}[!ht!] \centering
\includegraphics[width=0.65\textwidth]{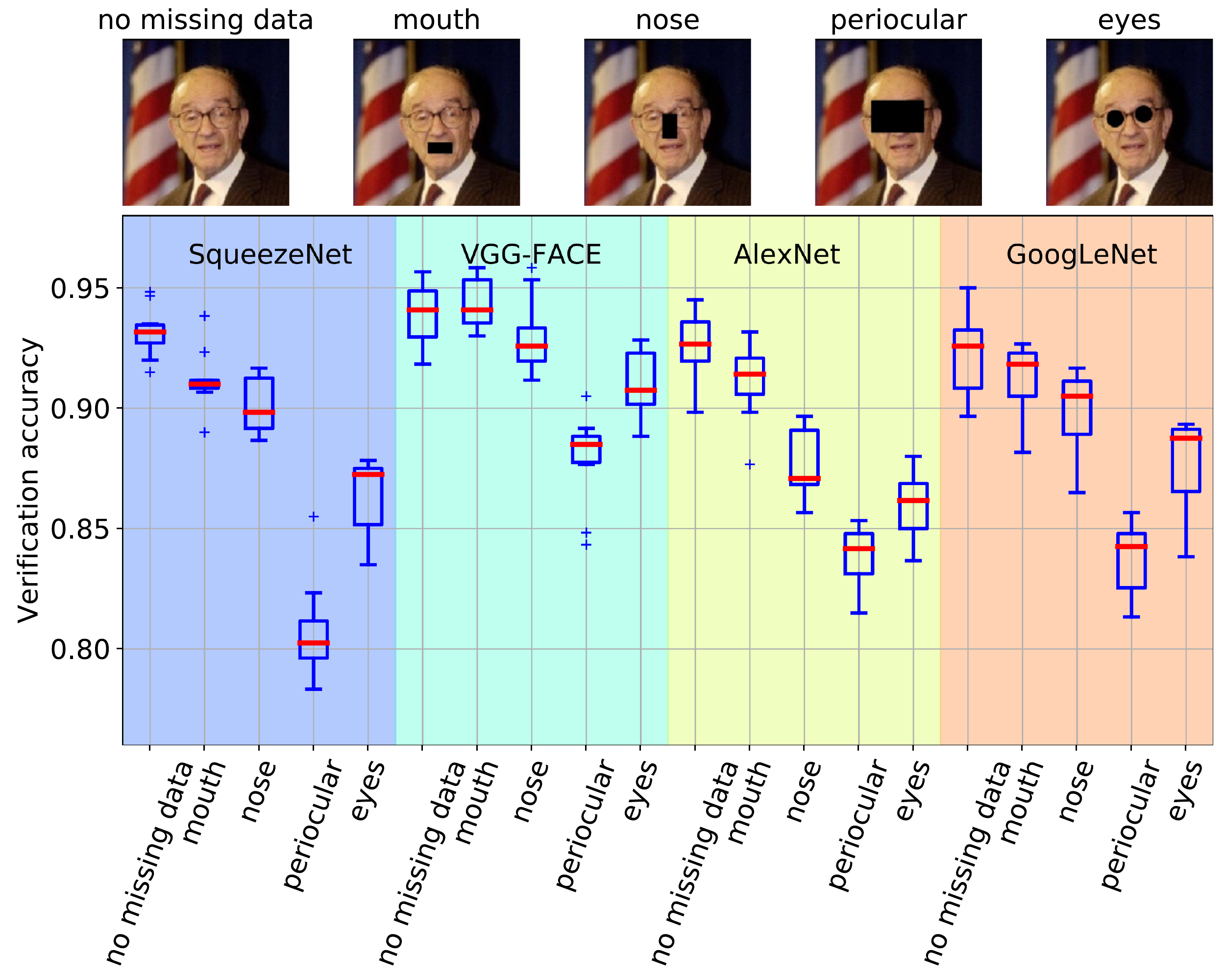}
\caption{Impact of missing data on the performance of the four deep models. The box plots show results for missing data at four different image locations, i.e., around the mouth, the nose, the periocular region and around the eyes. The box plots were computed from the $10$ experimental fold defined by the LFW verification protocol.}
\label{occlusions_plot}
\end{figure*}

In the first series of verification experiments, we explore the impact of Gaussian blur and JPEG compression. As can be seen in Fig.~\ref{LFW_crop_jpg} (left), image blurring has a significant effect on the performance of all deep models, which causes a quick drop in performance with an increase in the standard deviation of the Gaussian filters. Interestingly, the GoogLeNet model loses verification accuracy faster than the other three models. When looking at the impact of JPEG compression in Fig.~\ref{LFW_crop_jpg} (right), we see that all models are mostly unaffected by the compression artifacts until the compression quality is at its lowest possible value. Here, a compression quality of 0 corresponds to the scenario where all AC DCT coefficients are rounded to 0. Thus, only the DC components remain unaltered, and consequently every MCU is represented by a constant color. This is equivalent to uniformly downscaling the image by a factor of 8.
We observe that the verification accuracy of all models at the lowest JPEG quality roughly corresponds to the accuracy on the target images degraded with Gaussian blur with $\sigma = 5$, which is consistent with the above interpretation of the JPEG compression process in the sense that the amount of information preserved in the blurred and compressed images is approximately the same.

In the second series of experiments, we investigate the impact of Gaussian and salt-and-paper noise on the verification performance of the four deep models. From the results in Fig.~\ref{LFW_gaussian_blur_noise}, we see that the models behave similarly for both types of noise. The VGG-Face model performs the best and more robustly, followed by the AlexNet and SqueezeNet models, which perform more or less the same, and the GoogLeNet model, which is affected the most by the presence of noise. These results suggest that noise is an important factor affecting the performance of deep models and consequently that sufficiently low levels of noise need to be assured for reliable verification performance.   

In the third series of verification experiments, we study the effects of brightness and contrast.  
We can see from the results in Fig. \ref{LFW_brightness_contrast} (left) that the increase in brightness has a significant impact on the verification performance of the deep models and affects all models to more or less the same extent. In relative terms no model has an edge over the others  even at higher brightness factors, which is expected as important discriminative information is lost during the brightening process due to the pixel truncation. However, in absolute terms the VGG-Face model is the top performer ensuring the highest verification accuracy at all brightness factors. When looking at the results for different contrast factors in Fig. \ref{LFW_brightness_contrast} (right), we see that the relative performance of all models degrades similarly as the contrast decreases. In general the models are not particularly affected by the loss of contrast, as the verification accuracy remains well above $0.9$ even when more than $60\%$ of the contrast is removed. 

In the last series of experiments pertaining to image quality, we evaluate the effects of missing data on the verification performance. The results are displayed in Fig.~\ref{occlusions_plot} in the form of box plots. We can see that the impact of missing data follows the same relative ranking for all models: missing information around the periocular region is the most detrimental for the verification performance, followed in order by the eye, nose, and mouth regions. Interestingly, we can see that the VGG-Face model is the most affected by missing data around the periocular region, whereas the performance degradation for other regions is equal or lower than the degradations of the other models. We can also notice that the relative ranking of the tested models changes with respect to the image region, from which textural information was removed. While VGG-Face is the top performer in terms of average verification accuracy on the original images, it falls behind SqueezeNet and GoogLeNet when data around the eye, nose or periocular regions is missing. All in all, GoogLeNet appears to be the most robust to missing data, as the performance variations are the smallest with this model.

Overall, our experiments suggest that image quality is a crucial factor for the performance of existing deep models and that quality assessment of the input images should be an integral part of face recognition approaches based on deep learning. To mitigate problems pertaining to image quality image enhancement techniques need to be used or suitable data augmentation approaches need to be integrated into the training procedures to make the models robust against image-quality degradations.

\subsection{Impact of model-related covariates} \label{SubSec: model-covs}
\begin{figure*}[tb] \centering
\begin{minipage}{0.82\textwidth}
\begin{minipage}{0.48\textwidth}
\centering
\includegraphics[width=1\textwidth]{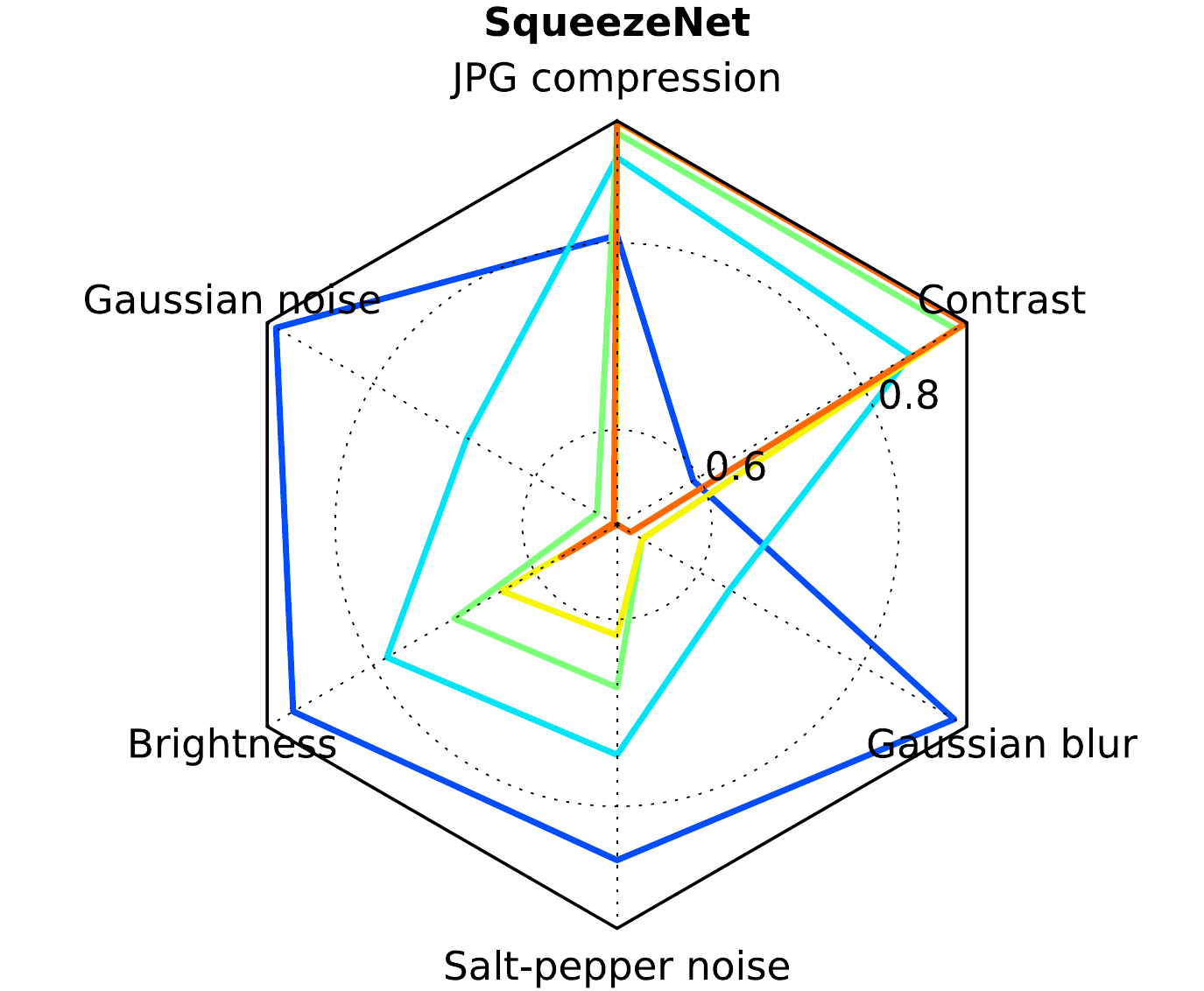}
\end{minipage}
\hfill
\begin{minipage}{0.48\textwidth}
\centering
\includegraphics[width=1\textwidth]{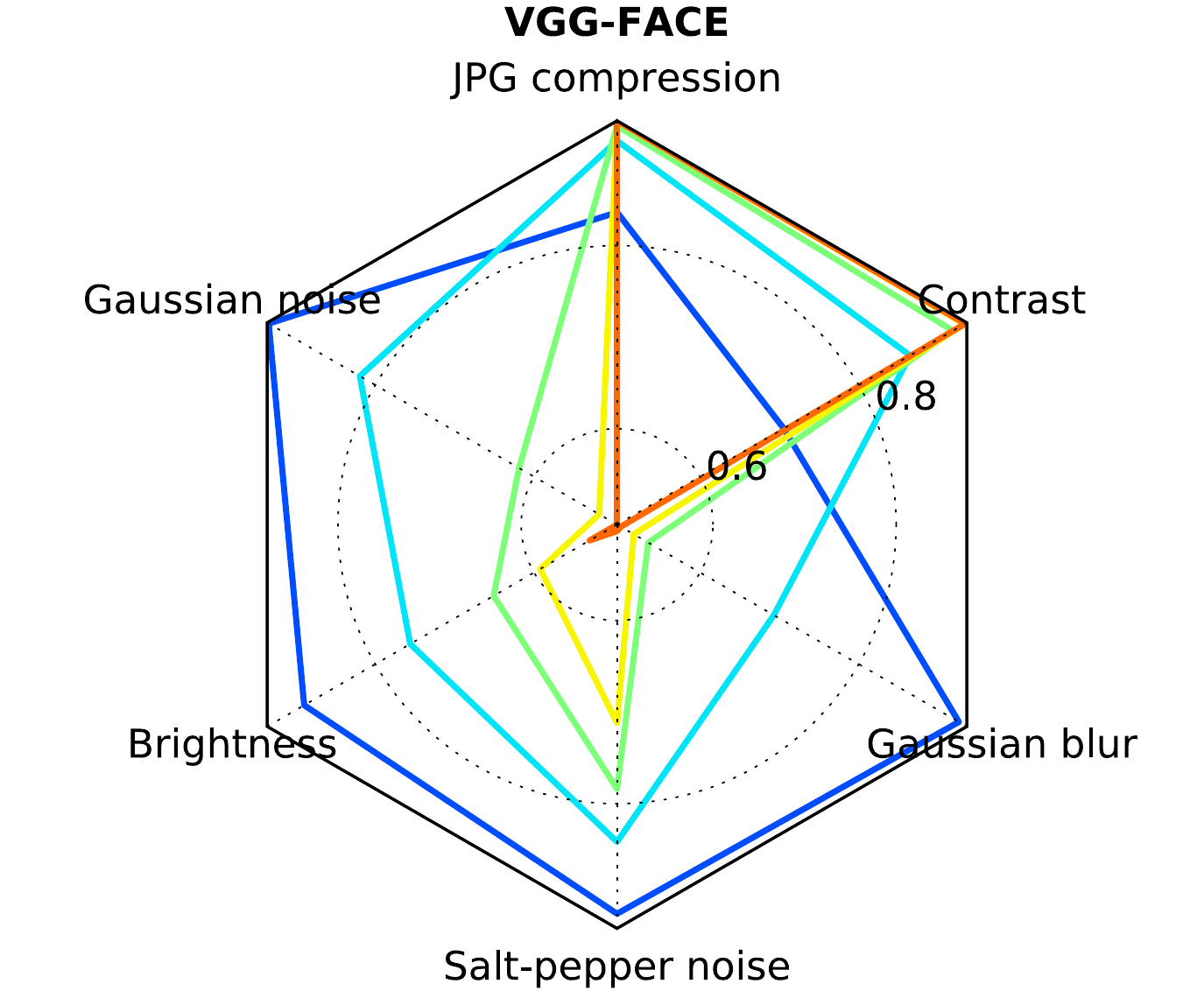}
\end{minipage}
\end{minipage}

\vspace{3mm}

\begin{minipage}{0.82\textwidth}
\begin{minipage}{0.48\textwidth}
\centering
\includegraphics[width=1\textwidth]{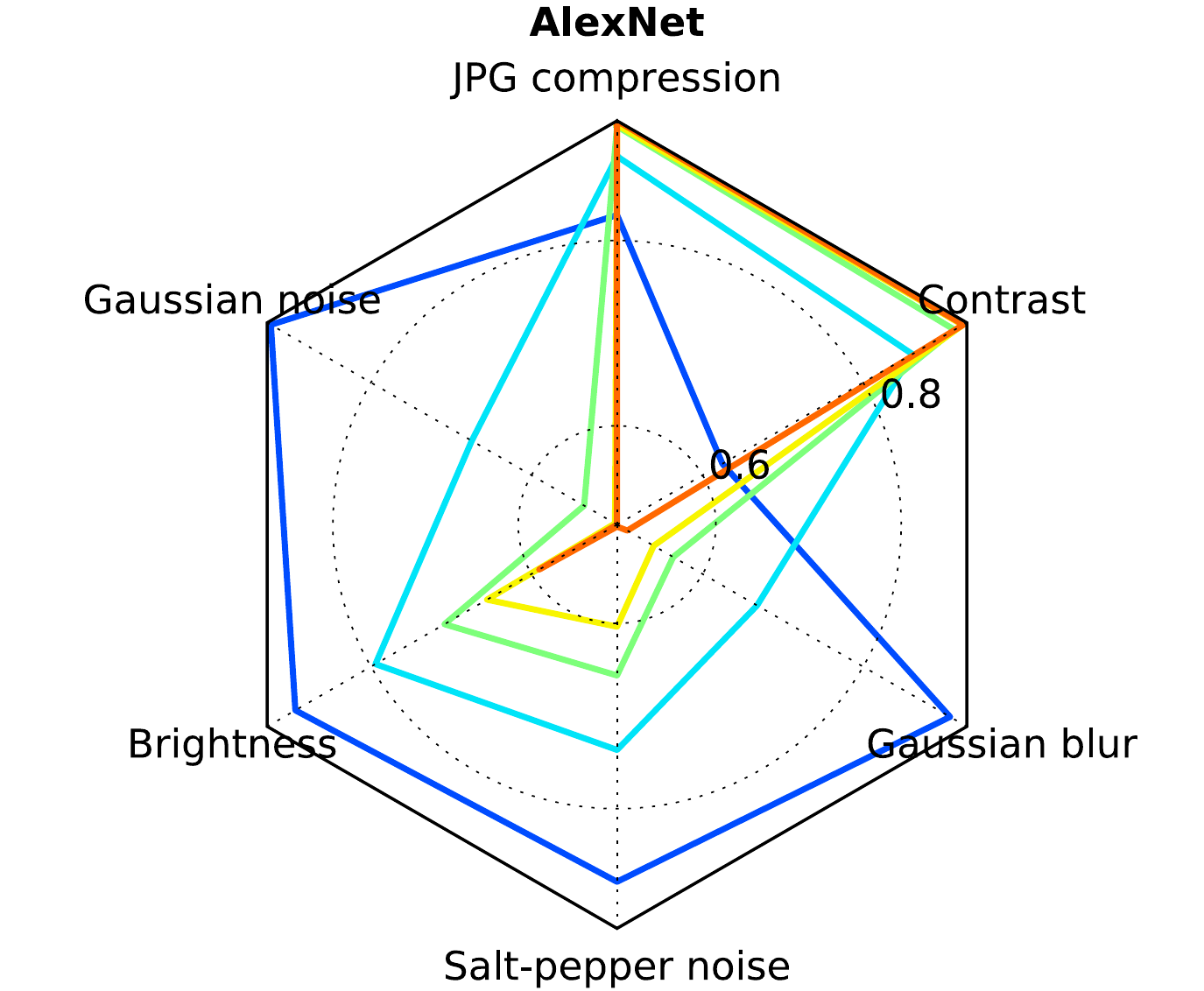}
\end{minipage}
\hfill
\begin{minipage}{0.48\textwidth}
\centering
\includegraphics[width=1\textwidth]{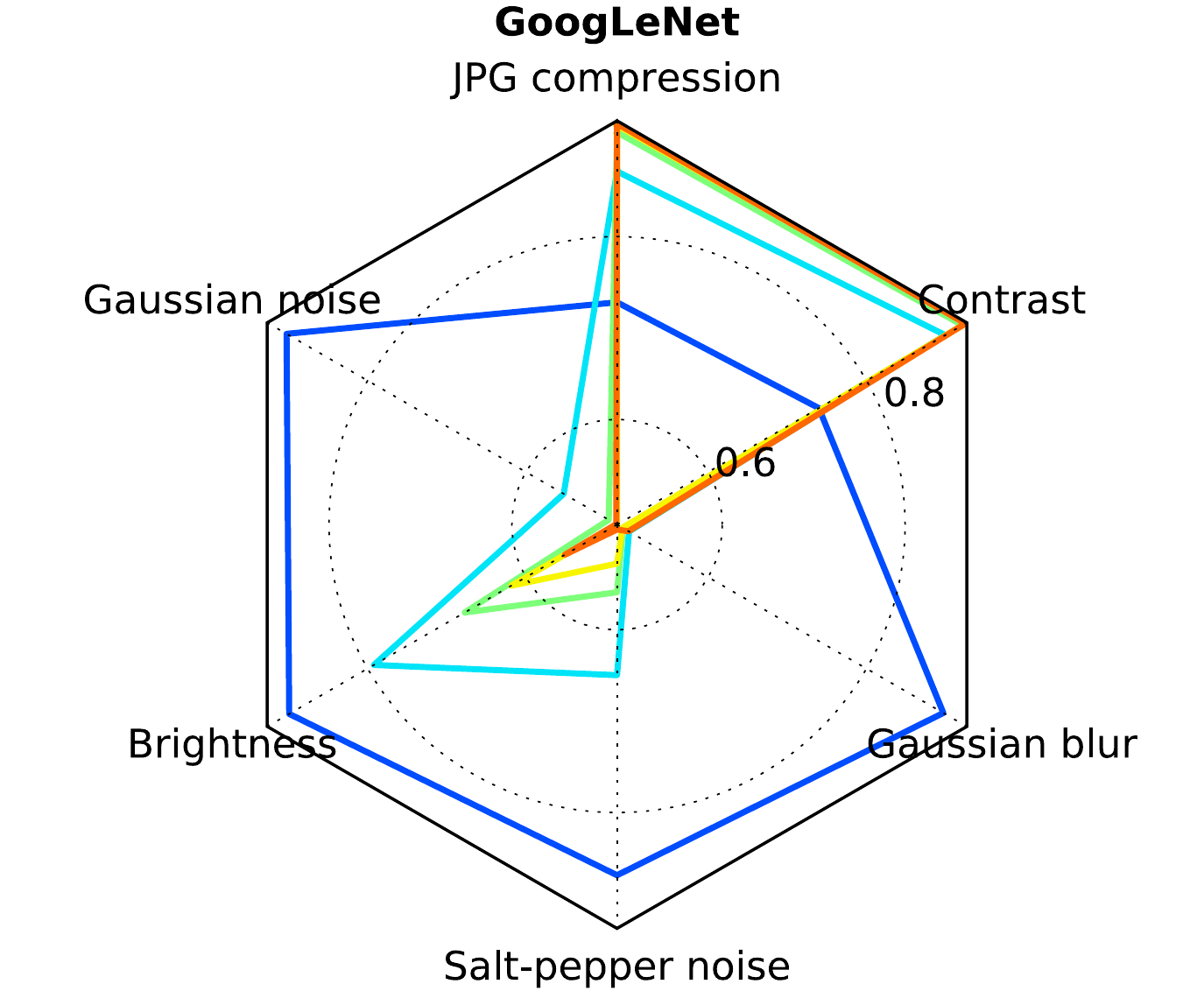}
\end{minipage}
\end{minipage}
\caption{Impact of the model architecture on the performance and robustness of the verification procedure. Here, the line colors correspond to the color-coded sample images on top of Figs.~\ref{LFW_crop_jpg}-~\ref{LFW_brightness_contrast}. A larger area covered by a curve indicates a better performance. The closer the curves of different color are in a given graph, the more robust the model is to image-quality degradations.}
\label{radar-charts}
\end{figure*}
\begin{figure}[t]
\centering{\includegraphics[width=1\columnwidth]{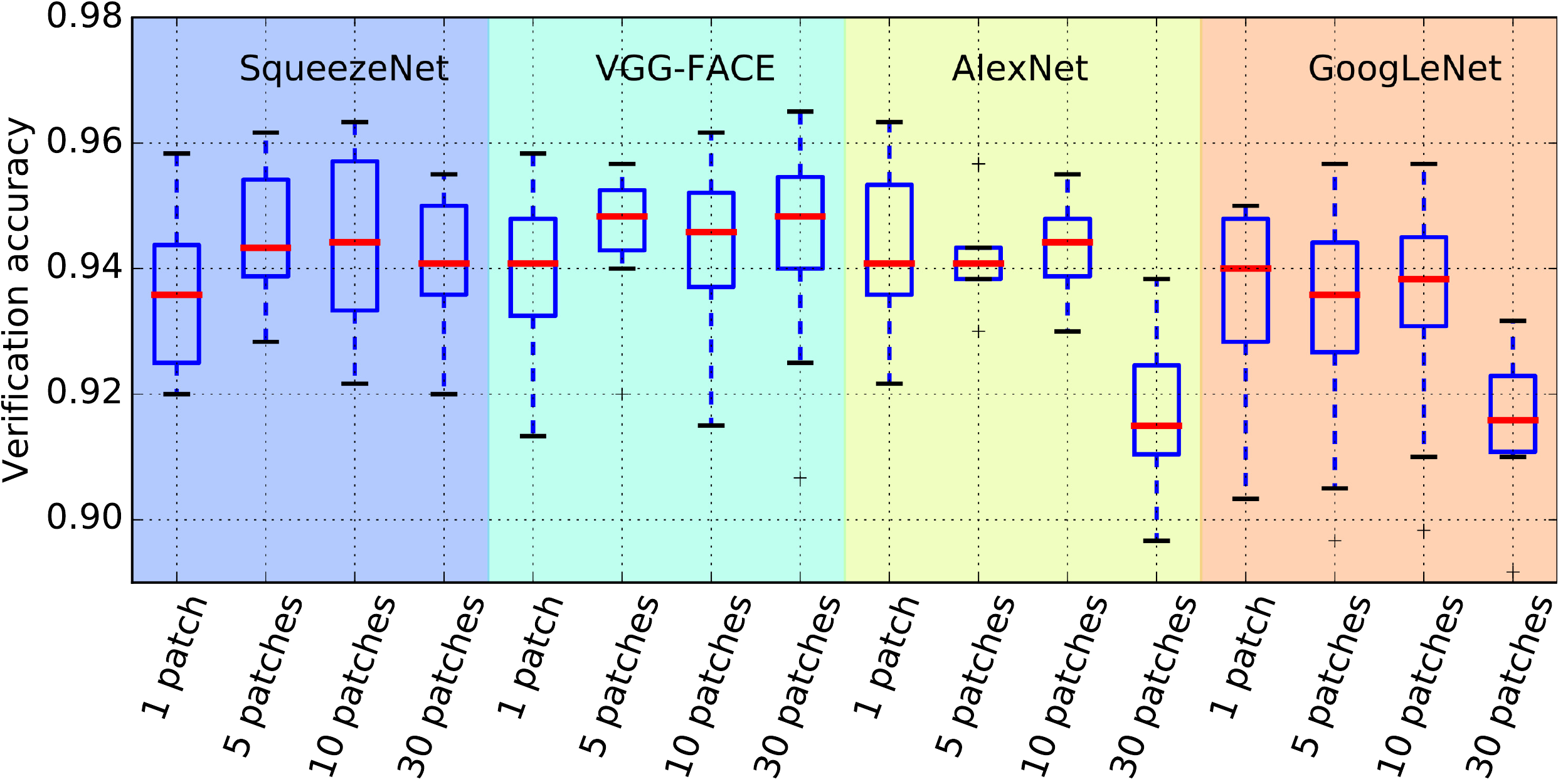}}
\caption{Performance evaluation for different sampling schemes. The box plots show results for the four sampling strategies, where image descriptors are computed based on either $1$, $5$, $10$ or $30$ face patches. The box plots were computed from the $10$ experimental fold defined by the LFW verification protocol.}
\label{subcrops_boxplot}
\end{figure}
\begin{figure}[t]
\centering{\includegraphics[width=1\columnwidth]{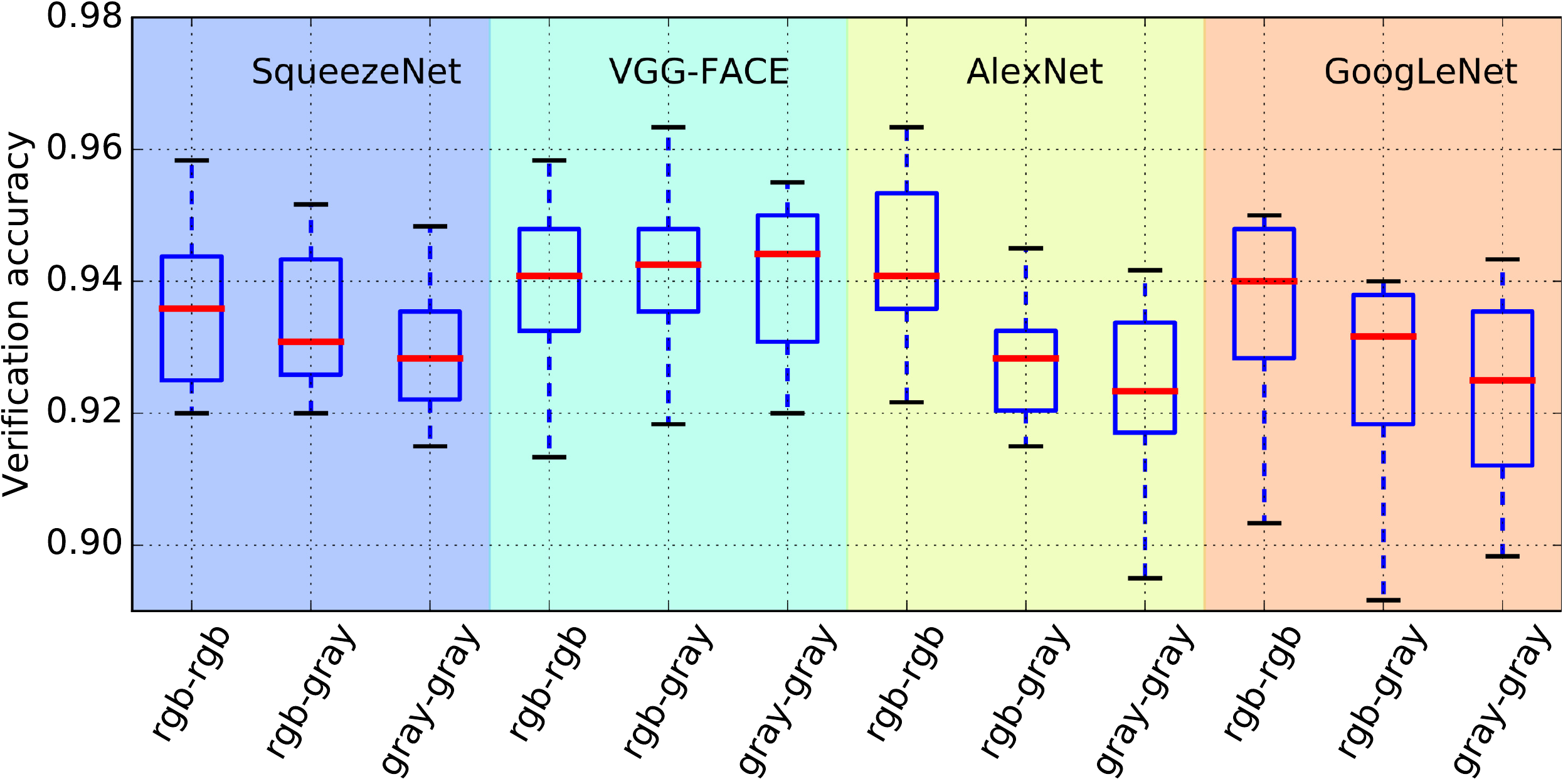}}
\caption{Performance evaluation between color and grayscale images. For each model, three comparisons were made: color-color, color-grayscale and grayscale-grayscale}
\label{rgb-gray}
\end{figure}
In the first series of experiments pertaining to model-related covariates, we assess the impact of different model architectures on the performance and robustness of the  LFW verification task. We present our comparison in the form of radar charts for different probe sets that correspond to the color-coded sample images at the top of Figs.~\ref{LFW_crop_jpg}-~\ref{LFW_brightness_contrast}. For example, the red curve in each chart corresponds to experiments with the probe images marked red in Figs.~\ref{LFW_crop_jpg}-~\ref{LFW_brightness_contrast}, the green curve to experiments with probe images marked green and so on.  Here, the larger the area covered by a curve the better the performance of the models across various image-quality covariates and the closer the different color curves are to each other for a given architecture, the robuster the architecture is to variations of the covariates. While all models perform similarly, the VGG-Face model has overall a slight advantage in term of robustness over the remaining three models. The SqueezeNet and AlexNet models perform almost the same, whereas our implementation of GoogLeNet is the least robust.

In the second series of verification experiments of this part, we evaluate the four different descriptor-computation strategies. The results of our experiments are presented in Fig. \ref{subcrops_boxplot} in the form of box-and-whiskers plots computed from the $10$ experimental folds defined by the LFW verification protocol. In these experiments we use the original LFW images without any degradations. We find that the SqueezeNet and VGG-Face models benefit marginally from averaging of the generated patch representations. While the trend shows an increase of $1$\%-$2$\% in verification accuracy by using more than a single patch to generate the image descriptors, the differences in performance are not statistically significant. The AlexNet and GoogLeNet models, on the other hand, do not show any improvements in performance. These results are unexpected as all models were trained with random patches sampled from the base face regions. We also note that while the SqueezeNet and VGG-Face models show some improvement when using 5 patches compared to only the central patch, there is no further improvement from the $10$- or $30$-patch schemes.

In the last series of experiments on model-related covariates we explore the impact of color information. The results of the experiment are shown in Fig. \ref{rgb-gray} again in the form of box plots. All models exhibit the best performance, when target and probe images are both in color, which is expected given that they were trained on color data exclusively. However, with the exception of AlexNet, we note that the accuracy of the models drops only marginally, when either the probe or both the target and probe images are switched to gray-scale. The difference in performance is not statistically significant, which points to a potential degree of redundancy in the models' architecture, observing that  eliminating two-thirds of the input information results in nearly identical performance.

\section{Conclusion}\label{Sec: conclusion}

We have presented a systematic study of covariate effects on face verification performance of four recent deep CNN models. We observe that the  studied models are affected by image quality to different degrees, but all of them degrade in performance quickly and significantly, when evaluated on lower-quality images than they were trained with. However, given proper architecture choices and training procedures, a deep learning model can be made relatively robust to common sources of image quality degradations. 

We found that the models considered were the most easily and consistently degraded in performance through image blurring, which is similar in nature to  real-life scenarios of attempting face recognition from low-resolution imagery. Other covariates found to have a considerable effect on the verification performance were noise, image brightness, and missing data, while image contrast and JPEG compression impacted the performance of the models only marginally.

Most of the models considered were least affected by changes in input color space - despite being trained on full color images -- their performance drops negligibly when evaluated on grayscale images. This finding is also corroborated by the results of the contrast  experiments. 

No specific architecture was found to be significantly more robust than others to all covariates. The VGG-Face model, for example, was most robust to noise, but performed least well for changes in image brightness. GoogLeNet, on the other hand, performed worst on noise and image blur, but had a slight advantage over the remaining models with images of reduced contrast.  

Based on our results, we identify the following prospective directions of further research related to deep models:
\begin{itemize}
	\item Image enhancement - various algorithms exist to enhance the appearance of blurred or low-resolution images for human perception. Given the low face recognition performance on such images, their applicability to automated face recognition systems is likely to be an important research direction for deep face recognition models in the future.
    \item Exploitation of color information - given the fact that most of the models we studied retained almost unaltered performance when presented with grey-scale images, it appears to be the case that the architectures considered here do not make proper use of color information in their input images. It follows that better deep learning models could be developed that either make more efficient use of their input information, or that discard color information altogether in favor of more compact models.
    \item Recognition from partial data - missing data proved to be a challenge for all evaluated models with performance deteriorating more, when larger contiguous areas of the images known to be important for identity inference were removed, e.g., the periocular region. This observation suggests that research into deep CNN models capable of recognition from partially observed data is needed and should be a focus of future research efforts. 
\end{itemize}

\section*{Acknowledgement}

This research was supported in parts by the ARRS (Slovenian Research Agency) Research Programme P2-0250 (B) Metrology and Biometric Systems, by ARRS through the junior researcher programme and by a Marie Curie FP7 Integration Grant within the 7th EU Framework Programme.

\bibliographystyle{ieee}
\bibliography{refs}

\end{document}